%% file: main.tex
\documentclass{article}
\usepackage{fullpage}

\input packages.tex
\input macros.tex

\usepackage{subfig}

\input{math_notation_for_this_paper}
\graphicspath{ {./figures/} }

\usepackage{pifont}

\usepackage{chngcntr}
\usepackage{adjustbox}
\usepackage{longtable}

\begin{document}


\title{
LoRA-Enhanced Distillation on Guided Diffusion Models
}

\author{
Pareesa Ameneh Golnari
\\  Microsoft \\ 
{\tt \small\{pagolnar\}@microsoft.com}
}

\date{}
\maketitle


\input{_s0_abstract.tex}
\input{_s1_intro.tex}
\input{_s2_proposed}
\input{_s3_experiments}

\input{_s4_conclusion}

\section*{Acknowledgements}
I would like to extend my appreciation to Zhewei Yao for inspiring ideas and insightful discussions that greatly influenced this project.
I also acknowledge Yuxiong He for her support and constructive comments, which added valuable perspectives to the project.


{
\bibliographystyle{plain}
\bibliography{ref.bib}
}

\clearpage

\end{document}

%% file: packages.tex
\usepackage[utf8]{inputenc} 
\usepackage[T1]{fontenc}    
\usepackage{url}            
\usepackage{booktabs}       
\usepackage{amsfonts}       
\usepackage{nicefrac}       
\usepackage{microtype}      
\usepackage{soul}
\usepackage{graphicx}
\usepackage{amsmath}
\usepackage{outlines}
\usepackage{xurl}

\usepackage[colorlinks=true,
citecolor=blue,
filecolor=black,
linkcolor=blue,
urlcolor=blue]{hyperref}

\input{math_commands.tex}

\usepackage{multirow}
\usepackage{paralist}

\usepackage{multicol}
\usepackage{diagbox}

\usepackage[ruled,noend]{algorithm2e}

\SetCommentSty{mycommfont}

\usepackage{here}

\usepackage{amsmath,amssymb,amsfonts,amsbsy,amsfonts,latexsym}
\usepackage{makecell}
\usepackage{xcolor}
\usepackage{colortbl}

\usepackage{tabularx,colortbl,xcolor}
\usepackage[normalem]{ulem}
\useunder{\uline}{\ul}{}

\usepackage{enumitem}

\usepackage{xparse}

\SetKwInput{KwInput}{Input}
\SetKwInput{KwRequire}{Require}

\usepackage{longtable}

%% file: math_commands.tex
\usepackage{amsmath,amsfonts,bm}

\usepackage{xcolor}
\usepackage{colortbl}

\def\eqref#1{equation~\ref{#1}}

\DeclareMathAlphabet{\mathsfit}{\encodingdefault}{\sfdefault}{m}{sl}
\SetMathAlphabet{\mathsfit}{bold}{\encodingdefault}{\sfdefault}{bx}{n}



%% file: macros.tex
\NewDocumentCommand{\var}{O{s} m O{}}{%
  \ensuremath{#1_{#2}^{#3}}
}
\usepackage{siunitx}

\newcommand{\commentout}[1]{}

\definecolor{light-gray}{gray}{0.80}

\usepackage{amsthm}

\usepackage{amsthm}

\newtheorem{mytheorem}{Theorem}
\newtheorem{assumption}[mytheorem]{Assumption}
\newtheorem{lemma}[mytheorem]{Lemma}
\newtheorem{remk}[mytheorem]{Remark}

%% file: _s0_abstract.tex
\begin{abstract}

Diffusion models, such as Stable Diffusion (SD)~\cite{SD}, offer the ability to generate high-resolution images with diverse features, but they come at a significant computational and memory cost. In classifier-free guided diffusion models, prolonged inference times are attributed to the necessity of computing two separate diffusion models at each denoising step.
Recent work has shown promise in improving inference time through distillation techniques, teaching the model to perform similar denoising steps with reduced computations~\cite{Distillation}. However, the application of distillation introduces additional memory overhead to these already resource-intensive diffusion models, making it less practical.

To address these challenges, our research explores a novel approach that combines Low-Rank Adaptation (LoRA)~\cite{Lora} with model distillation to efficiently compress diffusion models. This approach not only reduces inference time but also mitigates memory overhead, and notably decreases memory consumption even before applying distillation. The results are remarkable, featuring a significant reduction in inference time due to the distillation process and a substantial 50\% reduction in memory consumption.
Our examination of the generated images underscores that the incorporation of LoRA-enhanced distillation maintains image quality and alignment with the provided prompts. In summary, while conventional distillation tends to increase memory consumption, LoRA-enhanced distillation offers optimization without any trade-offs or compromises in quality.

\end{abstract}

%% file: _s1_intro.tex
\section{Introduction}
\label{sec:intro}

Diffusion models, exemplified by cutting-edge architectures like Stable Diffusion (SD)~\cite{SD}, GLIDE~\cite{GLIDE}, DALLE~\cite{DALLE}, and Imagen~\cite{Imagen}, have emerged as state-of-the-art generative models, enabling the generation of high-resolution images with a wide array of diverse features. However, these remarkable models come with inherent limitations, notably their substantial size, memory footprint, and prolonged inference times.

In the context of addressing the issue of extended inference times we previously demonstrated that certain iterations of denoising exhibit lower sensitivity, offering an opportunity for significant optimization~\cite{selective}. Recent advancements in the field have shown the potential to extend this optimization across all iterations through the use of distillation techniques~\cite{Distillation}. Yet, it is important to note that while distillation effectively mitigates inference time challenges, it introduces a new predicament by exacerbating memory consumption. This is primarily due to the necessity of storing the parameters of both the teacher and student models.

This paper introduces an innovative and synergistic approach that combines distillation with the concept of Low-Rank Adoptation (LoRA)~\cite{Lora}. This approach accomplishes two crucial objectives: first, it minimizes inference time, and second, it completely eliminates the additional memory overhead associated with distillation. Moreover, it significantly reduces the total memory consumption of the model even before applying distillation. This dual benefit not only optimizes the model's inference speed but also enhances its memory efficiency from the outset. Consequently, this novel technique holds the promise of advancing diffusion models, substantially enhancing their overall efficiency and resource utilization.

\subsection{Classifier-free diffusion models}

Classifier-free guidance is a valuable technique employed in diffusion models like GLIDE, Stable Diffusion, DALL·E 2, and Imagen to improve sample quality based on specific conditions. 
It introduces a guidance weight parameter \textit{"s"} to balance quality and diversity in generated samples. The approach involves generating both conditional and unconditional terms at each update step, with the model prediction computed using a weighted combination: 
\begin{equation}
    \label{guided-diffusion}
        \hat{\epsilon_\theta}(x_t|y)=\epsilon_\theta(x_t|0)+s.(\epsilon_\theta(x_t|y)-\epsilon_\theta(x_t|0)).
\end{equation}
Here $\epsilon_\theta(x_t|y)$ is the combined noise, $\epsilon_\theta(x_t|0)$ is the unconditional term and $\epsilon_\theta(x_t|y)$ is the conditional term and \textit{s} scales the perturbation.

\subsection{Applying Distillation to SD} 

Leveraging distillation presents an advantageous opportunity for optimizing guided diffusion models like SD. In~\cite{Distillation} the authors apply a two-stage distillation process to SD, resulting in a noticeable reduction in inference cost. For the sake of simplicity, we concentrate on the application of distillation to a single direction to demonstrate how we can effectively enhance efficiency while maintaining result quality. In this method, while the teacher model calculates the combined noise using two diffusion models [~\ref{guided-diffusion}], the student model computes only one diffusion model and, through the distillation process, learns to approximate the teacher's combined noise computation. This results in a remarkable 40\% improvement in inference time.  

This approach results in higher memory usage due to the need for duplicating weight matrices, driven by discrepancies between the teacher's and student's weights. The comparison of these weight matrices will be thoroughly explored in Section~\ref{sec:exp}. In the subsequent section, we introduce the application of LoRA in conjunction with distillation to address and mitigate this memory overhead issue.

%% file: _s2_proposed.tex
\section{Proposing LoRA-Enhanced Distillation}

\begin{figure}
  \centering
  \includegraphics[width=1.05\textwidth]{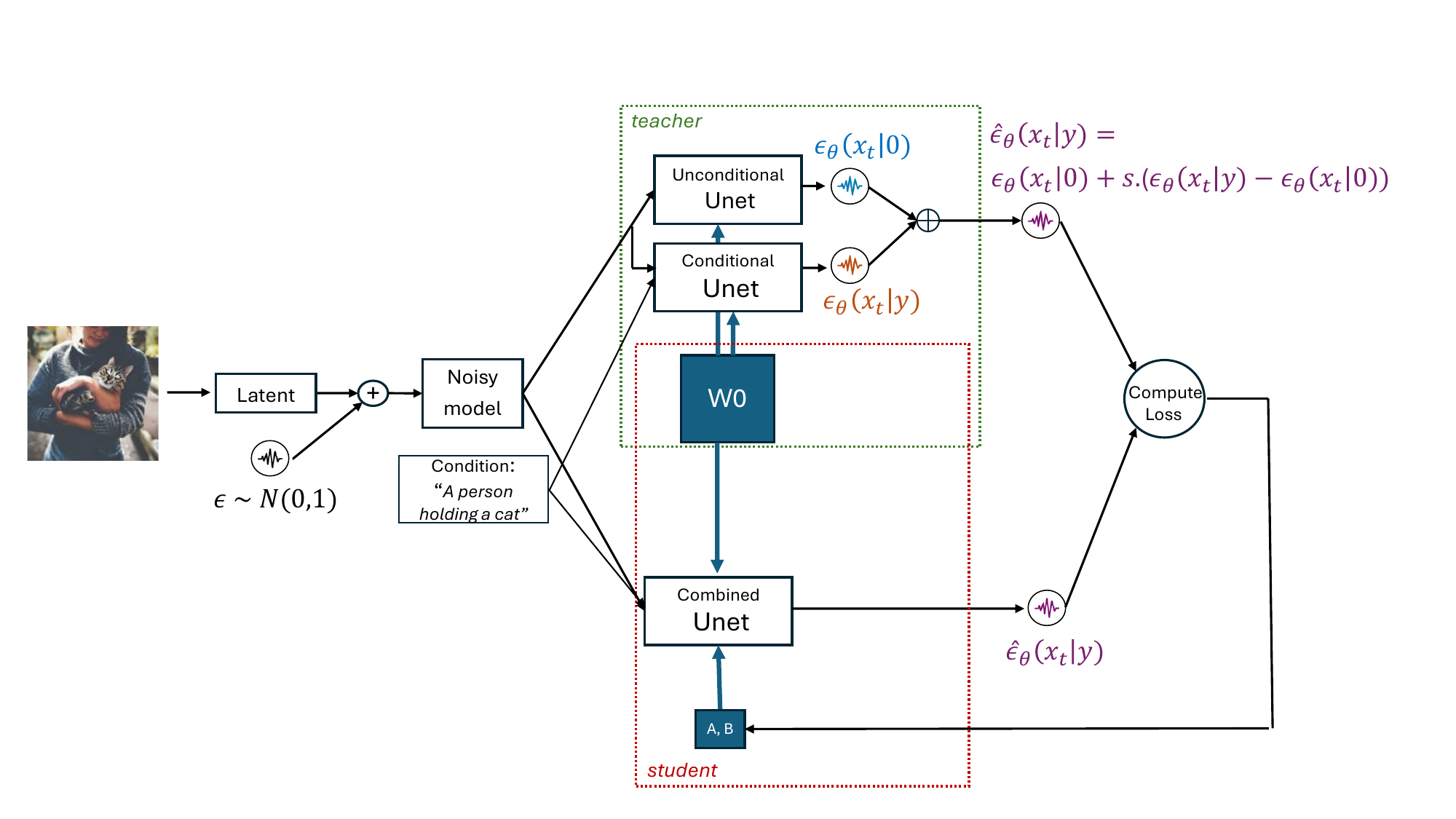}
  \caption{Diagram illustrating the proposed architecture in which the student (lower box) and the teacher (upper box) jointly use the $W_0$ matrix, with the training process concentrating on the low-rank matrices $A$ and $B$.}
  \label{diagram}
\end{figure}

In our approach, we implement the Low-Rank Adaptation (LoRA) method within the context of the distillation process, as illustrated in Figure \ref{diagram}. As shown in the diagram, both the teacher and the student models receive the noisy image in the latent space, along with additional conditions, such as prompts in text-to-image generation applications, for denoising. In this design, the teacher leverages the original guided diffusion architecture to compute two separate noises: conditional ($\epsilon_\theta(x_t|y)$) and unconditional noise ($\epsilon_\theta(x_t|0)$), and generates the combined noise ($\hat{\epsilon_\theta}(x_t|y)$) as referenced in \ref{guided-diffusion}. Through the distillation process, the student learns to compute the same noise with only half the computation, leading to a significant reduction in inference time \cite{Distillation}.

Regarding the weight handling, we adopt the LoRA technique \cite{Lora} to decompose the weight matrix into two components: the original weights (\textit{W$_0$}) and a set of low-rank matrices (\textit{A, B}). This decomposition empowers the model to retain crucial information in \textit{W$_0$} while efficiently updating and adapting the model's behavior using the lower-dimensional \textit{A, B} matrices. 

A noteworthy innovation in our approach is the shared use of the weight matrix \textit{W$_0$} between the student and the teacher. This approach entirely eliminates the need for separate memory resources dedicated to the teacher model. By sharing the same weight matrix, the teacher's information becomes inherently embedded within the student, negating the requirement for additional memory resources for the teacher.

Consequently, the integration of LoRA with distillation provides a dual advantage. Firstly, it completely eradicates the memory overhead associated with maintaining a separate teacher model, making it particularly valuable in scenarios where memory constraints are a significant concern. Secondly, LoRA's inherent ability to efficiently adapt and fine-tune neural network weights further reduces memory requirements during the adaptation process. Thus, this combined approach not only eliminates the need for teacher memory but also optimizes memory efficiency in weight adaptation, making it a valuable asset in resource-constrained machine learning settings like diffusion models.

%% file: _s3_experiments.tex
\section{Experiments}
\label{sec:exp}
In our experiments, we employed a guided diffusion model, the SD architecture ~\cite{SD}, and implemented it using the Huggingface pipeline~\cite{HF-SD} and Dreambooth~\cite{dreambooth}. Our training data consisted of a substantial dataset of pre-generated images by SD~\cite{DSdata}. These experiments were conducted on a Tesla V100 GPU running CUDA version 11.6.
We applied the LoRA technique on top of the distillation process for this model. Our results demonstrated that, while maintaining image quality, we achieved a 40\% reduction in inference time due to the distillation process and a 50\% reduction in memory consumption, highlighting the advantages of employing LoRA in this context.

\subsection{Image quality} 

\begin{figure}
  \centering
  \includegraphics[width=1.05\textwidth]{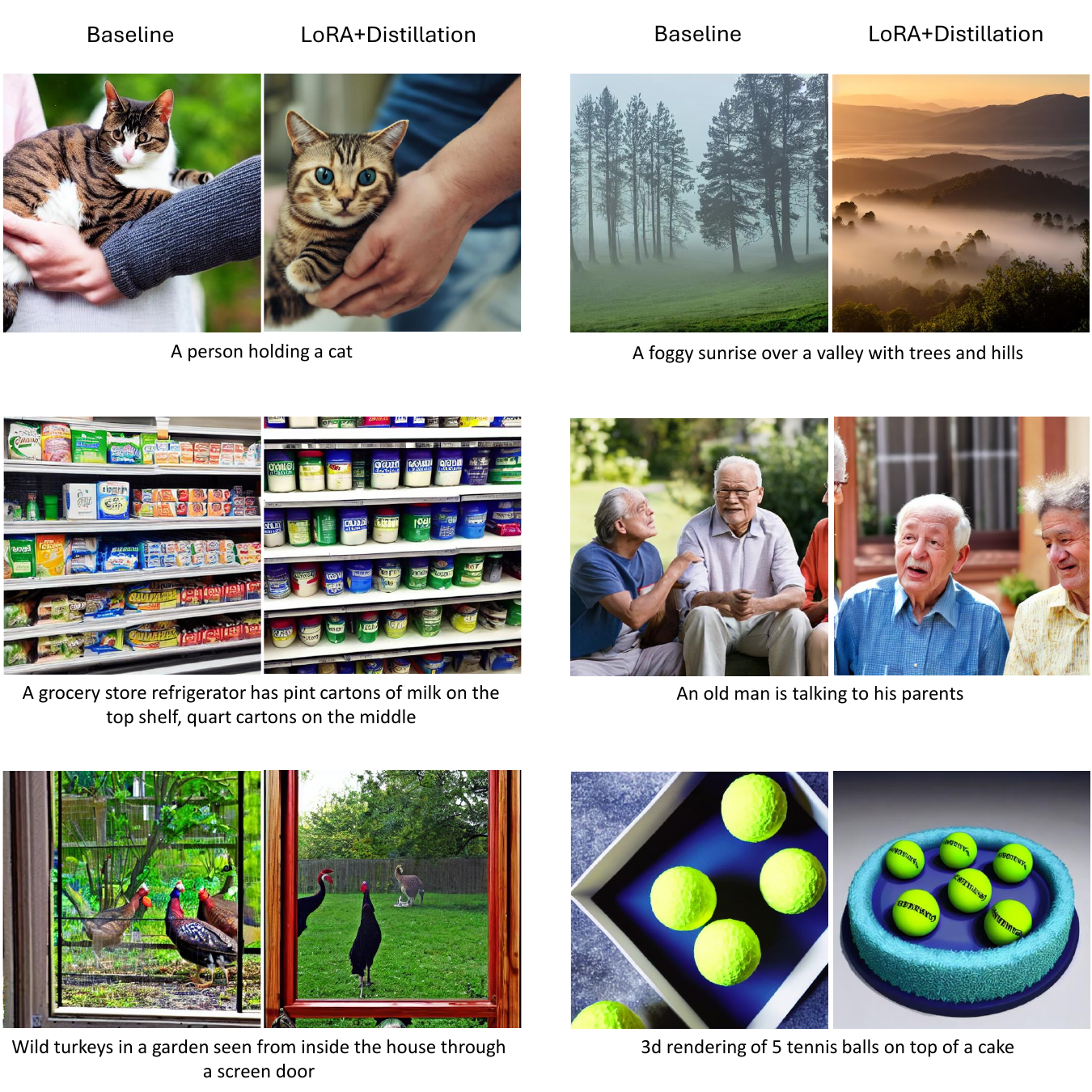}
  \caption{Contrasting the images generated by the baseline model (on the left-hand side) with those produced after applying LoRA-enhanced distillation (on the right-hand side)}
  \label{image-comparison}
\end{figure}

Figure \ref{image-comparison} illustrates a comparison of image quality. In the first column, you can observe baseline images, while the second column displays images generated with the application of distillation, keeping all other parameters constant, including the prompt and seed. As depicted in the figure, the utilization of LoRA-enhanced distillation does not degrade image quality nor compromise its alignment with the given prompt. In essence, while using distillation alone may undermine result quality, employing LoRA-enhanced distillation ensures optimization without introducing any trade-offs or compromises in terms of result quality.

\subsection{Memory consumption} 

\begin{table}[!h]
\begin{tabular}{|l|c|c|}
\hline
Model & Mem. Consumption & Memory Saving \\
\hline
Baseline SD & 21.0 GB & - \\
\hline
Baseline SD + Distillation & 24.4 GB & -16.2\% \\
\hline
Baseline SD + LoRA & 9.6 GB & 54.2\% \\
\hline
Baseline SD + Distillation + LoRA & 10.3 GB & 51.0\% \\
\hline
\end{tabular} 
\caption{The average measured time to generate an image using SD inference pipeline on Tesla V100 running CUDA version 11.6 and denoising iteration set to 50.}
\label{table-perf}
\end{table}

In Table \ref{table-perf}, we examine the maximum runtime memory consumption for each of the configurations. The "Memory Saving" column represents the ratio of the difference in memory consumption between the baseline model and the respective model to the memory consumption of the baseline model itself. As indicated in the table, applying distillation leads to a 16\% increase in memory consumption. However, when employing LoRA-enhanced distillation, not only is there no additional memory overhead, but there is also a significant improvement in memory savings, with an increase of 51\%.

%% file: _s4_conclusion.tex
\section{Conclusion}
In conclusion, our study presents an innovative approach to model distillation by incorporating the LoRA technique, effectively addressing the challenges associated with inference time reduction and memory consumption. Through empirical experiments, we have demonstrated significant improvements in both aspects. The integration of LoRA-enhanced distillation not only led to a remarkable reduction in inference time, showcasing the efficacy of distillation but also delivered an impressive 50\% reduction in memory consumption, eliminating the residual memory overhead typically associated with distillation. Moreover, our analysis of generated images underscores the robustness of this approach as it maintains image quality and alignment with prompts, even surpassing the results achieved with distillation alone. This work highlights the potential of LoRA-enhanced distillation as a powerful optimization technique for large diffusion models, offering unequivocal benefits in terms of efficiency and result quality.
\label{sec:conclusions}